\documentclass[10pt,twocolumn,letterpaper]{article}

\usepackage{cvpr}              

\makeatletter
\renewcommand{\paragraph}{%
  \@startsection{paragraph}{4}%
  {\z@}{1ex \@plus 1ex \@minus .2ex}{-1em}%
  {\normalfont\normalsize\bfseries}%
}
\makeatother
\usepackage{graphicx}
\usepackage{times}
\usepackage{amsmath}
\usepackage{colortbl}
\usepackage{amssymb}
\usepackage{xcolor}

\usepackage{booktabs}
\usepackage{bm}
\usepackage{pifont}
\usepackage{bbding}
\usepackage{amsbsy}
\usepackage{multirow}

\usepackage[pagebackref,breaklinks,colorlinks]{hyperref}

\usepackage[capitalize]{cleveref}
\crefname{section}{Sec.}{Secs.}
\Crefname{section}{Section}{Sections}
\Crefname{table}{Table}{Tables}
\crefname{table}{Tab.}{Tabs.}
\definecolor{topcolor}{rgb}{0.9020, 0.9608,0.9882}
\definecolor{bottomcolor}{rgb}{0.9804,0.94117647,0.91764706}

\definecolor{ugreen}{cmyk}{1,0,1,0.498}
\definecolor{lyyblue}{cmyk}{0.8278,0.3333,0,0.2941}
\definecolor{lyygreen}{cmyk}{0.6813,0,0.725,0.3725}
\definecolor{lyyred}{cmyk}{0,0.8855,0.8767,0.1098}
\definecolor{dblue}{cmyk}{1,0.5487,0,0.5569}
\definecolor{lightgrey}{gray}{0.90} 
\definecolor{lypurple}{HTML}{e0c2c0}
\definecolor{lygreen}{HTML}{eff67b}
\definecolor{lyblue}{HTML}{d5ddef}
\definecolor{lyyellow}{HTML}{fdfab5}
\definecolor{lypink}{HTML}{ffe0e5}
\definecolor{lyred}{HTML}{b71a3b}
\definecolor{lygrey}{HTML}{c4c0c2}

\def\cvprPaperID{3676} 
\def\confName{CVPR}
\def\confYear{2023}

\begin{document}

\title{GaitGCI: Generative Counterfactual Intervention for Gait Recognition}

\author{Huanzhang Dou$^1$ \quad Pengyi Zhang$^1$ \quad Wei Su$^1$ \quad Yunlong Yu$^2$\quad Yining Lin$^3$\quad Xi Li$^{1,4,5,6}$\thanks{Corresponding author.}\\
$^1$College of Computer Science \& Technology, Zhejiang University\\
$^2$College of Information Science \& Electronic Engineering, Zhejiang University\\
$^3$SupreMind \quad
$^4$Shanghai Institute for Advanced Study, Zhejiang University\\
$^5$Shanghai AI Laboratory\quad $^6$Zhejiang – Singapore Innovation and AI Joint Research Lab
}


\maketitle

\begin{abstract}
Gait is one of the most promising biometrics that aims to identify pedestrians from their walking patterns. However, prevailing methods are susceptible to confounders, resulting in the networks hardly focusing on the regions that reflect effective walking patterns. To address this fundamental problem in gait recognition, we propose a \textbf{G}enerative \textbf{C}ounterfactual \textbf{I}ntervention framework, dubbed GaitGCI, consisting of \textbf{C}ounterfactual \textbf{I}ntervention \textbf{L}earning (CIL) and \textbf{D}iversity-\textbf{C}onstrained \textbf{D}ynamic \textbf{C}onvolution (DCDC). CIL eliminates the impacts of confounders by maximizing the likelihood difference between factual/counterfactual attention while DCDC adaptively generates sample-wise factual/counterfactual attention to efficiently perceive the sample-wise properties. With matrix decomposition and diversity constraint, DCDC guarantees the model to be efficient and effective. Extensive experiments indicate that proposed GaitGCI: 1) could effectively focus on the discriminative and interpretable regions that reflect gait pattern; 2) is model-agnostic and could be plugged into existing models to improve performance with nearly no extra cost; 3) efficiently achieves state-of-the-art performance on arbitrary scenarios (in-the-lab and in-the-wild).

\end{abstract}
\section{Introduction}
Gait recognition aims to utilize walking patterns to identify pedestrians without explicit cooperation, thus drawing rising attention. Current gait recognition research focuses on in-the-lab~\cite{1699873,takemura2018multi} and in-the-wild scenarios~\cite{zhu2021gait,zhang2023large} for theoretical analysis and practical application, respectively.
 \begin{figure}[t]
 \centering
    \begin{center}
       \includegraphics[width=0.465\textwidth]{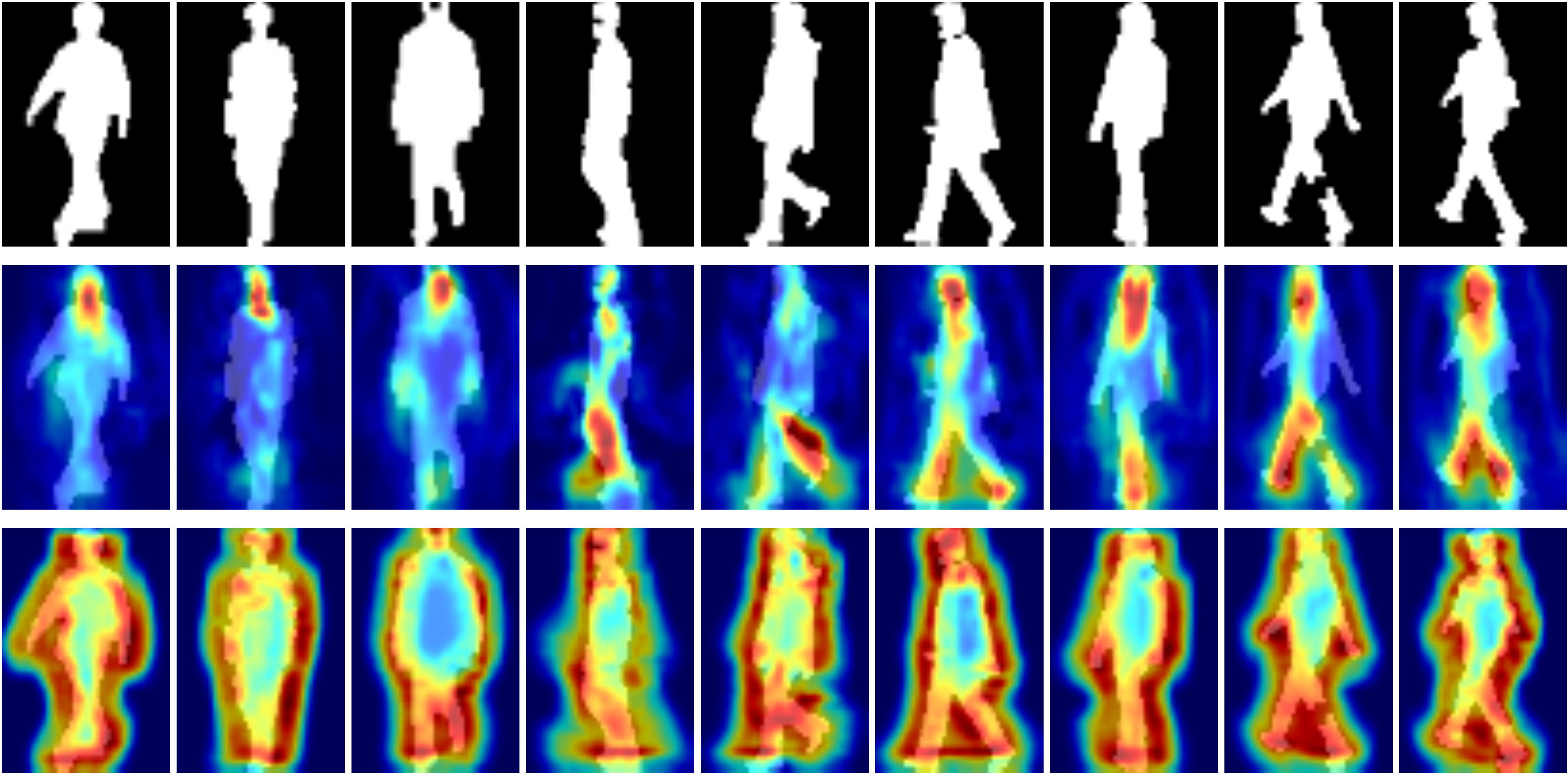}
    \end{center}
    \caption{Network attention comparison. From top to down: silhouette, existing method, and proposed GaitGCI. The confounders make the existing model collapse into suboptimal attention regions. By contrast, GaitGCI could effectively focus on the discriminative and interpretable regions (\ie., close to the boundary~\cite{liang2022gaitedge,wang2002gait}) that could represent walking patterns.}
    \label{fig:gradcam}
 \end{figure}

The key to addressing gait recognition is to fully capture the effective visual cues of the gait patterns, \ie., the regions close to the body boundary~\cite{liang2022gaitedge,wang2002gait} for both in-the-lab scenarios and in-the-wild scenarios. However, the attention analysis~\cite{8954073,Bai_2022_CVPR,wang2010chrono} on prevailing methods in~\cref{fig:gradcam} indicates that the existing methods hardly capture the effective gait patterns and tend to collapse into the suboptimal attention regions, which would deteriorate the gait representation. 
We argue that this phenomenon is caused by the network's susceptibility to the \textit{confounders}~\cite{geirhos2020shortcut,izmailov2022feature}, which may provide \textit{shortcuts}~\cite{geirhos2020shortcut,izmailov2022feature} for the models rather than the valid gait-related patterns. For example, the attention regions of prevailing methods are related to viewpoints~\cite{9275377} or walking conditions~\cite{Huang_2021_ICCV}. As shown in~\cref{fig:gradcam}, the prevailing network tends to focus on the head under the front view and the head/feet under the side view. However, the majority of the gait-related information close to the boundary is neglected. Therefore, how to alleviate the impact of confounders is a fundamental problem to model discriminative and interpretable gait representation.

 \begin{figure}[t]
    \begin{center}
       \includegraphics[width=0.49\textwidth]{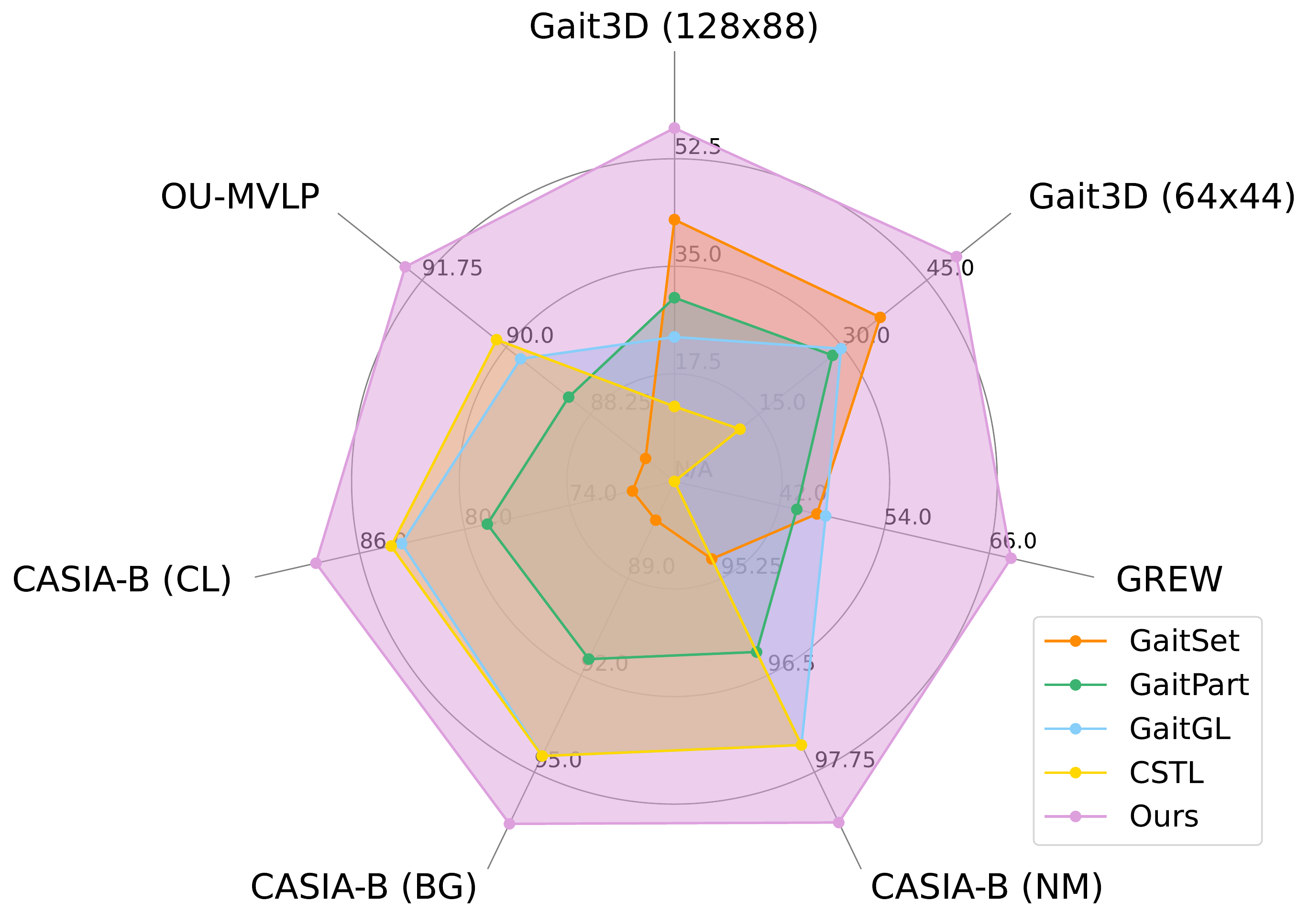}
    \end{center}
    \caption{GaitGCI could achieve state-of-the-art performance under arbitrary scenarios, including in-the-lab scenarios~\cite{1699873,takemura2018multi} and in-the-wild scenarios~\cite{Zheng_2022_CVPR,zhu2021gait}.}
    \label{fig:compare}
 \end{figure}

Motivated by this, we propose a generative counterfactual intervention framework, named GaitGCI, consisting of Counterfactual Intervention Learning (CIL) and Diversity-Constrained Dynamic Convolution (DCDC). The core idea of CIL is to leverage the counterfactual-based causal inference to alleviate the impact of confounders and mine the direct causality link between factual attention and prediction.
Specifically, we first construct a causal analysis tool (\ie., Structural Causal Model~\cite{pearl2000models})  to formulate the causality links among the input, attention, and prediction. Then, the training objective is modified from maximizing the original likelihood that contains confounders to maximizing the likelihood difference between the factual/counterfactual attention, which forces the network to focus on the direct causality between the factual attention and the prediction instead of collapsing into the confounders.

Further, considering that 
the previous network to produce factual attention is static and the mainstream counterfactual is pre-defined distribution~\cite{9710619,chang2021towards} (\eg., random or normal distribution), which limits the ability of the network to perceive the sample-wise properties. Therefore, we propose a Diversity-Constrained Dynamic Convolution (DCDC) to efficiently produce the sample-adaptive kernel, which aims to generate factual/counterfactual attention. Specifically, we first decouple the dynamic convolution~\cite{Verelst_2020_CVPR,NEURIPS2019_f2201f51} into the sample-agnostic convolution and sample-adaptive convolution. Then, to improve the efficiency, we apply the matrix decomposition to decompose sample-adaptive convolution into two bases and a generative affinity matrix, which transforms dense convolution integration in high-dimensional space into the aggregation of bases in low-dimensional space. Besides, to guarantee the representation power, we propose a rank-based diversity constraint on two bases of the sample-adaptive convolution.

By alleviating the impact of confounders, the proposed method: (1) could effectively focus on the discriminative and interpretable regions instead of collapsing into the confounders; (2) is model-agnostic and could boost the performance of prevailing methods; (3) could efficiently achieve state-of-the-art performance under arbitrary scenarios (in-the-lab and in-the-wild) as shown in~\cref{fig:compare}.

The main contributions are summarized as follows:
\begin{itemize}
\setlength{\itemsep}{0pt}
\setlength{\parsep}{0pt}
\setlength{\parskip}{0pt}
    \item We present counterfactual intervention learning (CIL) to alleviate the impact of confounders. CIL could effectively force the model to focus on the regions that reflect gait patterns by maximizing the likelihood difference between factual/counterfactual attention. 
    \item We present diversity-constrained dynamic convolution (DCDC) to generate factual/counterfactual attention in a sample adaptive manner. Matrix decomposition and diversity constraint guarantee efficiency and representation power, respectively.
    \item Extensive experiments demonstrate that the proposed framework efficiently achieves state-of-the-art performance in arbitrary scenarios. Besides, the proposed methods could serve as a plug-and-play module to boost the performance of prevailing models.

\end{itemize}

\section{Related Work}
\subsection{Gait Recognition}
Prior research focuses on the in-the-lab scenario. However, VersatileGait~\cite{zhang2023large} has pioneered the more challenging in-the-wild gait recognition via synthetic datasets. This problem draws increasing attention, resulting in the emergence of real-world datasets for in-the-wild scenarios~\cite{zhu2021gait,Zheng_2022_CVPR}. And mainstream methods could be grouped as follows:

\noindent\textbf{Silhouette-based Methods.} This fashion~\cite{Huang_2021_ICCV2,dou2022metagait} extracts gait patterns from the silhouette sequence. GaitSet~\cite{chao2019gaitset} deems each sequence as an unordered set, GaitPart~\cite{Fan_2020_CVPR} proposes part-based modeling, and GaitGL~\cite{Lin_2021_ICCV} extracts features from global/local representation. This paradigm is sensitive to covariates but is more popular for its efficiency.

\noindent\textbf{Skeleton-based Methods.} Many 
 methods~\cite{6117582,bodor2009view,4378964,1613073,liao2020model,kastaniotis2016pose,wang2004fusion} utilize pose estimation to model gait patterns. For example, Teepe~\textit{et al.}~\cite{9506717} model the skeleton as a graph and utilize GCN~\cite{kipf2017semi}. Li \textit{et al.}~\cite{li2020end} propose to jointly utilize 2D/3D keypoints information to model gait representation. These methods should be more robust to the covariates but rely on accurate pose estimation.

\noindent\textbf{Methods using Other Modalities.} Recently, more gait modalities have emerged. Several methods~\cite{li2020end, Zhang2022OnLD,liang2022gaitedge} extract features from RGB video. Castro~\textit{et. al.}~\cite{8053503} leverage optical flow to obtain abundant motion information. The depth information~\cite{nunes2019benchmark} and 3D mesh~\cite{li2020end,Zheng_2022_CVPR} are also introduced to use extra information. Further, several works~\cite{castro2020multimodal,hofmann2014tum,Zheng_2022_CVPR} conduct multi-modal learning to achieve informative representation.

 \begin{figure*}[ht]
    \begin{center}
       \includegraphics[width=0.95\textwidth]{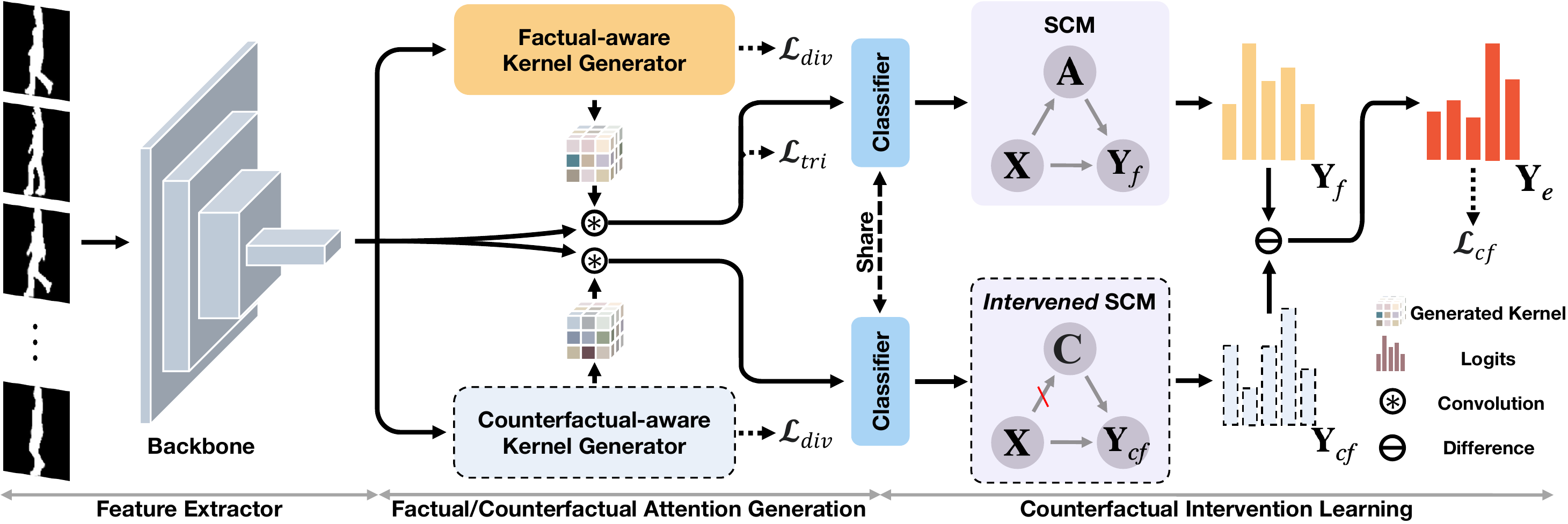}
    \end{center}
    \caption{Overview of GaitGCI. The factual/counterfactual-aware generator is implemented by the proposed diversity-constrained dynamic convolution to efficiently generate factual/counterfactual attention  based on the sample-wise properties. Then, counterfactual intervention learning is performed to maximize the likelihood difference between factual/counterfactual attention. The optimization objective is the combination of triplet loss, counterfactual loss, and diversity constraint.}
    \label{fig:overview}
 \end{figure*}

\subsection{Vision Causal Inference} Causal inference~\cite{Ramanishka_2018_CVPR,Fire_2017_CVPR_Workshops,wang2020visual} arouses widespread attention to endow networks with the ability to analyze the causal effect. The causal inference has been successfully used in various areas, including visual explanation~\cite{pmlr-v97-goyal19a,hendricks2016generating}, semantic segmentation~\cite{dong_2020_conta}, and few-shot/zero-shot learning~\cite{10.5555/3495724.3495954, Yue2021CounterfactualZA}. Previous vision causal inference methods with counterfactuals ~\cite{khorram2022cycle, abbasnejad2020counterfactual} focus on the analysis of the outcome intervened by sorts of pre-defined counterfactuals. By contrast, we leverage dynamic convolution to adaptively perceive the sample-wise factual/counterfactual attention.

\subsection{Dynamic Deep Neural Networks}  Dynamic network~\cite{Han2022DynamicNN,Su_2023_CVPR} aims to boost the network capacity and generalizability via adapting its parameters or structures based on the input during inference. Dynamic convolution~\cite{Verelst_2020_CVPR,NEURIPS2019_f2201f51} aggregates multiple candidate convolutions via the SE-style attention mechanism~\cite{hu2018squeeze}. DRConv ~\cite{Chen2021DynamicRC} proposes grouped dynamic convolution to adaptively select channels from groups. Besides, weight adjustment could be performed by soft attention over the spatial dimension of the convolutional weights~\cite{harley_segaware,su2019pixel,zhang2022adaptive}. In this paper, we propose to leverage matrix decomposition~\cite{li2021revisiting} and diversity constraint to guarantee the efficiency and representation power of dynamic convolution, respectively.

\section{Method}
\subsection{Overview}
As shown in~\cref{fig:overview}, the silhouette is first fed to the backbone with low-rank 3D CNN. Then, factual/counterfactual attention is generated by the corresponding kernel generator (diversity-constrained dynamic convolution). Finally, GaitGCI is optimized with counterfactual loss, triplet loss, and diversity constraint. The feature aggregation (temporal pooling/separate FC~\cite{chao2019gaitset}) is omitted for simplicity.

\subsection{Counterfactual Intervention Learning}
We propose Counterfactual Intervention Learning (CIL) to alleviate the impact of confounders. First, we formulate the learning process with the causality analysis tool, \ie, the Structural Causal Model (SCM)~\cite{pearl2000models,pawlowski2020deep}. Then, the counterfactual intervention is introduced to analyze the direct causality link between factual attention and prediction.

\noindent\textbf{Structural Causal Model Formulation.} To represent the causality links among input $\mathbf{X}$, attention $\mathbf{A}$, and prediction $\mathbf{Y}$, we formulate them with the SCM $\bm{\mathcal{G}}=\{\mathbf{N}, \mathbf{E}\}$, where $\mathbf{N}$ and $\mathbf{E}$ represent the variable nodes and causality links, respectively. The causality links denotes: \textit{cause}$\to$ \textit{effect}. Therefore, the causality could be formulated as $\mathbf{X}\to \mathbf{Y}$: the conventional model. $\mathbf{X}\to \mathbf{A}$: the model produces the corresponding attention. $\mathbf{X}\to \mathbf{Y}\gets \mathbf{A}$: the final prediction $\mathbf{Y}$ is determined by $(\mathbf{X}, \mathbf{A})$ jointly. With SCM, The causality links between the variables can be directly analyzed via variable intervention, which means manipulating the value of specific variables and then observing the effect.

\noindent\textbf{Counterfactual Intervention.} 
Ideally, $\mathbf{A}$ decides to predict $\mathbf{Y}$ entirely by sensing the effective properties of $\mathbf{X}$. However, there are confounders in $\mathbf{X}$, which confuses the network's learning process and makes the network collapse into the suboptimal attention regions. Therefore, we propose to leverage the counterfactual intervention $Do(\cdot)$, which could cut off the causality link between the confounders and the factual attention. 

 The counterfactual intervention $Do(\cdot)$ could remove the impact of specific variables. Note that counterfactual~\cite{kusner2017counterfactual,verma2020counterfactual} means ``counter to the facts," and the intervention is impossible to occur in the real world. Thus the process of $Do(\cdot)$ is called \textit{counterfactual intervention}, which is achieved by an imaginary intervention to replace the variables' state. For example, the value of the counterfactual intervention $Do(\mathbf{A}=C)$ means that the counterfactual $C$ is assigned to $\mathbf{A}$ and breaks the causality link between $\mathbf{A}$ and its all parent nodes, which forces the variable to no longer be affected by the confounders. Therefore, the direct causality link between the factual attention $\mathbf{A}$ and the prediction $\mathbf{Y}$ could be analyzed. Specifically, the value $A$ and $C$ of factual attention $\mathbf{A}$ and counterfactual attention $\mathbf{C}$ is produced by the process $\mathcal{A}(\cdot)$ and $\mathcal{C}(\cdot)$, respectively. 

\begin{equation}
    A = \mathcal{A}(\mathbf{X})= \{\mathbf{A}_0, \mathbf{A}_1,..., \mathbf{A}_{M-1}\},
\end{equation}
\begin{equation}
    C = \mathcal{C}(\mathbf{X})= \{\mathbf{C}_0, \mathbf{C}_1,..., \mathbf{C}_{M-1}\},
\end{equation}
where $M$ is the channel number of $A$ and $C$ to control the capacity to perceive the sample-wise properties. In prevailing implementations, $\mathcal{A}(\cdot)$ is a static network, and $\mathcal{C}(\cdot)$ is a manually pre-defined distribution (\eg., random or normal distribution). Then, the likelihood of counterfactual intervention $\mathbf{P}(\mathbf{Y}|Do(\mathbf{A}=C))$ could be leveraged to analyze the direct causality link between $\mathbf{A}$ and $\mathbf{Y}$ excluding the confounders. The likelihood of factual attention $\mathbf{Y}_{f}$ and counterfactual intervention $\mathbf{Y}_{cf}$ could be formulated as:
\begin{equation}
    \mathbf{Y}_{f}=\mathbf{P}(\mathbf{Y}|\mathbf{A}=A)=\mathbf{\mathbb{E}}_{A\sim \mathcal{A}(\mathbf{X})}(\mathbf{X}*A),
\end{equation}
\begin{equation}
    \mathbf{Y}_{cf}=\mathbf{P}(\mathbf{Y}|Do(\mathbf{A}=C))=\mathbf{\mathbb{E}}_{C\sim \mathcal{C}(\mathbf{X})}(\mathbf{X}*C).
\end{equation}
 
 The former $\mathbf{Y}_f$ is the key to model discriminative and interpretable gait representation with gait-related properties, and the latter $\mathbf{Y}_{cf}$ denotes the context-specific confounders, which is expected to be removed from the likelihood prediction.
 Then, we calculate the likelihood difference~\cite{pearl2001direct} between the factual attention and the counterfactual attention to obtain the direct causality effect $\mathbf{Y}_{e}$ between the factual attention $\mathbf{A}$ and the corresponding prediction $\mathbf{Y}$:
\begin{equation}
    \mathbf{Y}_e = \mathbf{Y}_{f} - \mathbf{Y}_{cf}.
\end{equation}

Maximizing the likelihood difference $\mathbf{Y}_{e}$ could force the network to focus on factual attention learning instead of collapsing into the confounders represented by the counterfactuals. Thus, counterfactuals can be regarded as additional supervision to alleviate the impact of confounders.

Note that CIL is model-agnostic and could be a plug-and-play module. Besides, the impact of confounders is a fundamental problem, thus CIL could theoretically be applied to arbitrary scenarios. Further, CIL is only used during training and is discarded at the inference stage.

\subsection{Diversity-Constrained Dynamic Convolution}
We propose Diversity-Constrained Dynamic Convolution (DCDC) to adaptively generate factual/counterfactual attention based on the following observations. First, the existing attention module is static, which hinders models from perceiving the sample-wise properties of the sparse silhouette. Second, previous counterfactuals are from pre-defined distribution, which cannot adaptively represent the confounders of specific samples.

\noindent\textbf{Vanilla Dynamic Convolution.} The main idea of dynamic convolution~\cite{Verelst_2020_CVPR,NEURIPS2019_f2201f51} $\mathbf{W}(\mathbf{X})$ is to linearly combine $S$ static candidate convolutions $\{\mathbf{W}_s\}$ through the score $\{\pi_s(\mathbf{X})\}$ adaptively produced by the SE-style attention~\cite{hu2018squeeze} as:
\begin{equation}
    \resizebox{.9\linewidth}{!}{$
            \displaystyle
\mathbf{W}(\mathbf{X})=\sum_{s=1}^S\pi_s(\mathbf{X})\mathbf{W}_s \;
\text{s.t.} \; 0 \leq \pi_s(\mathbf{X}) \leq 1, 
\sum_{s=1}^S \pi_s(\mathbf{X}) = 1.
        $}
        \label{eq:dynamic-conv}
\end{equation}
\noindent\textbf{Reformulation with Matrix Decomposition.} To avoid the high costs from the high-dimensional computation~\cite{chen2020dynamic}, we reformulate the dynamic convolution with matrix decomposition. First, each candidate convolution $\mathbf{W}_s$ could be re-defined as the combination of a sample-agnostic kernel $\mathbf{W}_0$ and the corresponding offset kernel $\Delta\mathbf{W}_s$, \ie., $\mathbf{W}_s = \mathbf{W}_0 + \Delta\mathbf{W}_s$, where $\mathbf{W}_0 = \frac{1}{S}\sum_{s=1}^S \mathbf{W}_s$. Thus, the dynamic convolution $\mathbf{W}(\mathbf{X})$ could be reformulated as:
\begin{equation}
    \begin{aligned}
\mathbf{W}(\mathbf{X})&=\sum_{s=1}^S \pi_s(\mathbf{X})\mathbf{W}_0 + \sum_{s=1}^S\pi_s(\mathbf{X})\Delta \mathbf{W}_s \\
&=\mathbf{W}_0 + \sum_{s=1}^S\pi_s(\mathbf{X})\Delta \mathbf{W}_s.
\end{aligned}
\end{equation}

Specifically, $\mathbf{W}_0$ and $\{\pi_s(\mathbf{X})\Delta \mathbf{W}_s\}$ could be regarded as the kernel to extract sample-agnostic features and sample-adaptive features, respectively. Further, we propose to leverage low-rank decomposition on the sample-adaptive kernel to improve the efficiency as follows:

 \begin{figure}[t]
    \begin{center}
       \includegraphics[width=0.47\textwidth]{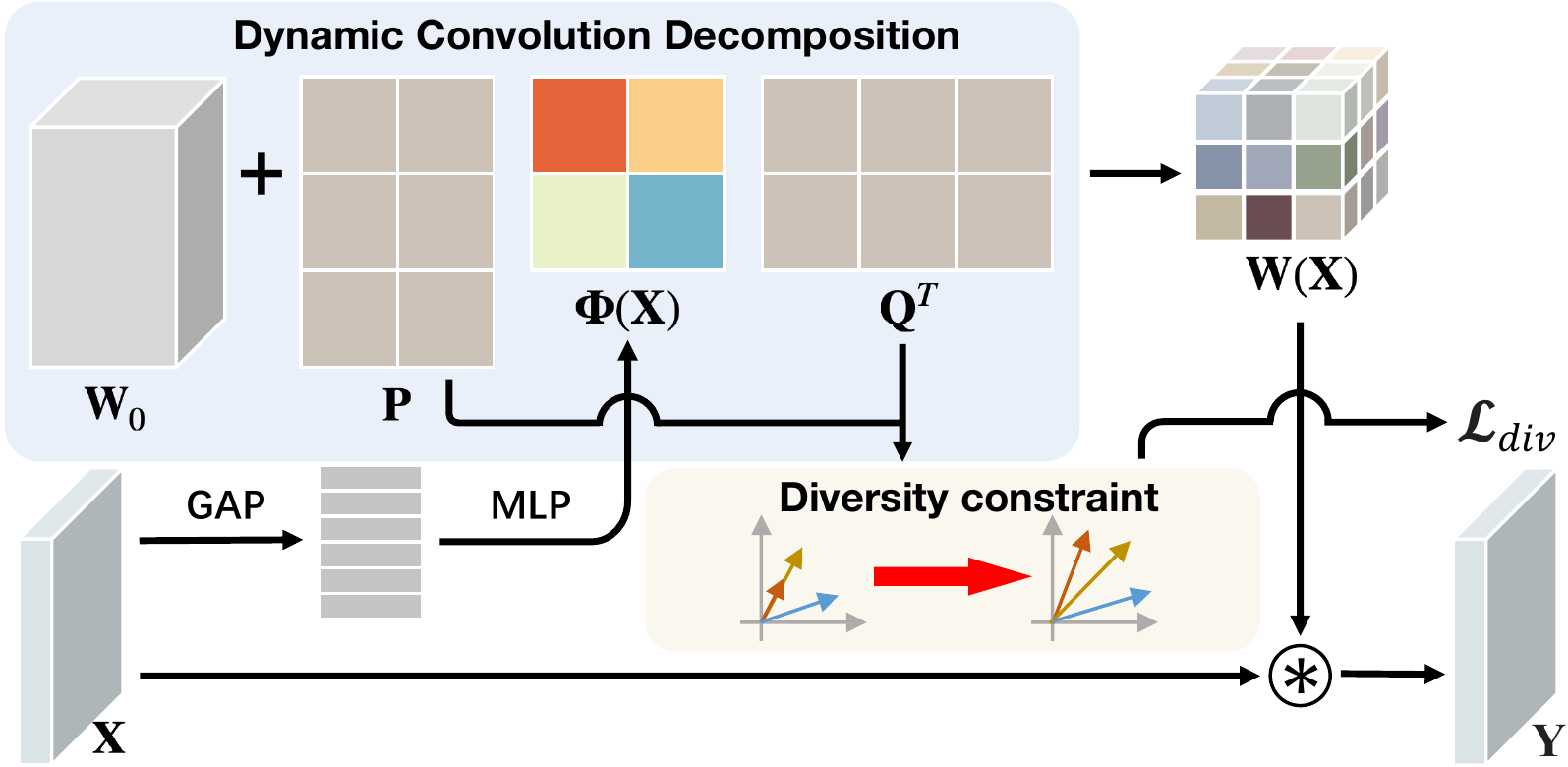}
    \end{center}
    \caption{Illustration of Diversity-Constrained Dynamic Convolution. DCDC is formulated as a sample-agnostic convolution $\mathbf{W}_0$ and sample-adaptive one, which could be decomposed into two bases $\mathbf{P}$/$\mathbf{Q}$ and an affinity matrix $\mathbf{\Phi(\mathbf{X})}$. Rank-based diversity constraint on two bases aims to guarantee the representation power.}
    \label{fig:kernel}
 \end{figure}

\begin{equation}
\begin{aligned}
\mathbf{W}(\mathbf{X})&=\mathbf{W}_0 + \sum_{i=1}^{L} \sum_{j=1}^{L} \mathbf{p}_i\phi_{i,j}(\mathbf{X})\mathbf{q}_j^T\\
&= \mathbf{W}_0 + \mathbf{P}\mathbf{\Phi}(\mathbf{X})\mathbf{Q}^T,
\label{eq:dynamic-residual}
\end{aligned}
\end{equation}
 where $\mathbf{P}\in\mathbb{R}^{C_{out}\times L}$ and $\mathbf{Q}\in \mathbb{R}^{k\times C_{in}\times L }$ are bases to interact the input in low-dimensional latent space $\mathbb{R}^L$. $k$ is the kernel size. $\mathbf{\Phi}(\cdot)\in \mathbb{R}^{L\times L}$ denotes  affinity matrix to adaptively interact $\mathbf{P}$ and $\mathbf{Q}$.  Therefore, the adaptiveness of dynamic convolution is transformed from the attention-based linear combination to the generative aggregation of two bases. And $\mathbf{\Phi (\mathbf{X})}$ could be generated by an MLP: 
 \begin{equation}
     \mathbf{\Phi(\mathbf{X})}= \mathbf{W}_{fc2}\times\delta({\mathbf{W}_{fc1}\times(GAP(\mathbf{X}))}),
 \end{equation}
 where $\mathbf{W}_{fc1}\in\mathbb{R}^{C/r\times C}$ and $\mathbf{W}_{fc2}\in\mathbb{R}^{L^2\times C/r}$. $\delta (\cdot)$ denotes the Sigmoid. In this way, the decomposition-base dynamic convolution could efficiently reduce the dimension of the latent space from $SC$ to $L$ ($SC\gg L$).

\noindent\textbf{Rank-based Diversity Constraint.} To guarantee the representation power, we propose to diversify two bases $\mathbf{P}$ and $\mathbf{Q}$. The diversity of the weight matrix $\mathbf{W}\in\mathbb{R}^{m\times n}$ could be represented by the rank function as:

\begin{equation}
   Rank(\mathbf{W})=\sigma_1^0+\sigma_2^0+\cdots +\sigma_r^0=\lim_{p\to 0}||\mathbf{W}||_{S_p}^p,
\end{equation}
where $\sigma_i$ is the $i^{th}$ singular value of the weight matrix $\mathbf{W}$ and $r=\mathrm{min}\{m,n\}$. The rank function $Rank(\cdot)$ has similar form with Schatten p-norm $||\cdot||_{S_p}^p$ ($p\to 0$)~\cite{tomioka2013convex, wang2016schatten}, which could be defined as: 
\begin{equation}
    ||\mathbf{W}||_{S_p}=(\sigma_1^p+\sigma^p_2+\cdots +\sigma_r^p)^{1/p} .
\end{equation}

However, optimizing rank is NP-hard and the Schatten p-norm ($p\neq 1$) is non-convex~\cite{xu2017unified}. Further, Schatten 1-norm (nuclear norm) $||\mathbf{W}||_{S_1}$ has been verified to be a convex approximation~\cite{lu2015nonconvex} to $Rank(\mathbf{W})$ and is differentiable as: 

\begin{equation}
     \frac{\partial||\mathbf{W}||_{S_1}}{\partial{\mathbf{W}}}=\frac{tr(\partial{\mathbf{\Sigma}})}{\partial{\mathbf{W}}}=\frac{tr(\mathbf{U}^T\partial(\mathbf{\mathbf{W}})\mathbf{V})}{\partial{\mathbf{W}}}=\mathbf{U}\mathbf{V}^T,
\end{equation}
where $\mathbf{W}$ is decomposed into $\mathbf{U}\mathbf{\Sigma}\mathbf{V}^T$ by singular value decomposition (SVD), which introduces nearly no extra computation since the representation is low-dimensional. Thus, we propose to leverage Schatten 1-norm as the diversity constraint to maximize $Rank(\mathbf{W})$:

\begin{equation}
    \mathcal{L}_{div} = -\sum_{i\in\{\mathbf{A},\mathbf{C}\}}||\mathbf{P}_i||_{S_1}-\sum_{j\in\{\mathbf{A},\mathbf{C}\}}||\mathbf{Q}_j||_{S_1}.
\end{equation}

~\label{sec:dcdc}
\subsection{Optimization}
To effectively optimize GaitGCI, the objective is composed of counterfactual loss $\mathcal{L}_{cf}$, triplet loss $\mathcal{L}_{tri}$~\cite{hermans2017defense}, and diversity constraint $\mathcal{L}_{div}$. Specifically, $\mathcal{L}_{cf}$ can be easily implemented with cross-entropy loss by replacing the original prediction $\mathbf{Y}$ with causality effect $\mathbf{Y}_{e}$.

\begin{equation}
     \mathcal{L}_{total} = \underbrace{\mathcal{L}_{ce}(\mathbf{Y}_{e},y)}_{Counterfactual\,\,Loss}+ \mathcal{L}_{tri} + \lambda\mathcal{L}_{div},
    \label{eq:mmcell}
\end{equation}
where $y$ is the ground truth and $\lambda$ is the weight of diversity constraint, respectively.

\section{Experiments}
\subsection{Dataset}

\noindent\textbf{OU-MVLP~\cite{takemura2018multi}.} It is one of the largest gait datasets, which includes 10307 subjects and each subject contains two sequences. The viewpoints are uniformly distributed between [$0^{\circ}$,$90^{\circ}$] and [$180^{\circ}$,$270^{\circ}$]. Following the mainstream protocol~\cite{chao2019gaitset}, the first sequence of each ID is deemed as the gallery, and the rest are the probe during the evaluation.

\noindent\textbf{CASIA-B~\cite{1699873}.} CASIA-B contains 124 subjects, and the viewpoints are distributed in [$0^{\circ}$, $180^{\circ}$]. Besides, 10 groups of three conditions are included in each subject, \ie., 6 normal (NM), 2 with a bag (BG), and 2 with a coat (CL). For evaluation, we adopt the mainstream protocol~\cite{chao2019gaitset}, which selects the first 74 subjects as the training set and the rest as the test set. During the evaluation, the sequences (NM\#01-NM\#04) are the gallery, and the rest are the probe.

\begin{table*}[!ht]
\centering
\small
\renewcommand{\arraystretch}{0.8}
\setlength{\tabcolsep}{1.6mm}
\caption{Rank-1 (\%) performance comparison on OU-MVLP, excluding the identical-view cases.}
\begin{tabular}{l|c|cccccccccccccc|c}
\toprule
\multirow{2}{*}{Method} & \multicolumn{1}{c|}{\multirow{2}{*}{Venue}} & \multicolumn{14}{c|}{Probe View}                                                               & \multirow{2}{*}{Mean} \\ \cmidrule{3-16}
                        & \multicolumn{1}{c|}{}                      & 0$^{\circ}$ & 15$^{\circ}$ & 30$^{\circ}$ & 45$^{\circ}$ & 60$^{\circ}$ & 75$^{\circ}$ & 90$^{\circ}$ & 180$^{\circ}$ & 195$^{\circ}$ & 210$^{\circ}$ & 225$^{\circ}$ & 240$^{\circ}$ & 255$^{\circ}$ & \multicolumn{1}{c|}{270$^{\circ}$} &      \\ \midrule
                                            GaitSet~\cite{chao2019gaitset}                                      & AAAI19  &79.5&87.9&89.9&90.2&88.1&88.7&87.8&81.7&86.7&89.0&89.3&87.2&87.8&86.2&87.1           \\
                      GaitPart~\cite{Fan_2020_CVPR}   &CVPR20                                  &   82.6&88.9&90.8&91.0&89.7&89.9&89.5&85.2&88.1&90.0&90.1&89.0&89.1&88.2&88.7    \\
                    GLN~\cite{hou2020gait}           &ECCV20                               &  83.8&90.0&91.0&91.2&90.3&90.0&89.4&85.3&89.1&90.5&90.6&89.6&89.3&88.5&89.2         \\
                    CSTL~\cite{Huang_2021_ICCV}              &ICCV21                           &   87.1&91.0&91.5&91.8&90.6&90.8&90.6&89.4&90.2&90.5&90.7&89.8&90.0&89.4&90.2           \\
                      3DLocal~\cite{Huang_2021_ICCV2}       &ICCV21                               &   86.1&91.2&92.6&\textbf{92.9}&92.2&91.3&91.1&86.9&90.8&\textbf{92.2}&92.3&91.3&91.1&90.2&90.9        \\
                     GaitGL~\cite{Lin_2021_ICCV}        &ICCV21                               &  84.9&90.2&91.1&91.5&91.1&90.8&90.3&88.5&88.6&90.3&90.4&89.6&89.5&88.8&89.7       \\
                     GaitMPL~\cite{9769988}        &TIP22                               &  83.9&90.1&91.3&91.5&91.2&90.6&90.1&85.3&89.3&90.7&90.7&90.7&89.8&88.9&89.6     \\
                     Lagrange~\cite{Chai_2022_CVPR}   &  CVPR22                                          & 85.9  & 90.6    & 91.3   & 91.5   & 91.2   & 91.0   &90.6    &88.9     &89.2     &90.5     &90.6     & 89.9    & 89.8    &     89.2                     & 90.0    \\ \midrule
                     \rowcolor{lightgrey}\textbf{GaitGCI}                                         & -- &\textbf{91.2}&\textbf{92.3}&\textbf{92.6}&92.7&\textbf{93.0}&\textbf{92.3}&\textbf{92.1}&\textbf{92.0}&\textbf{91.8}&91.9&\textbf{92.6}&\textbf{92.3}&\textbf{91.4}&\textbf{91.6}&\textbf{92.1}    \\ \bottomrule

\end{tabular}
 \label{tab:ou}
\end{table*}
\begin{table}[t]
\caption{Rank-1 (\%), parameters (M), and computation cost (G MACs) comparison at the inference stage on CASIA-B.}
\centering
\small
\renewcommand{\arraystretch}{0.8}
\setlength{\tabcolsep}{0.9mm}
\begin{tabular}{l|c|cccc|cc}
\toprule
Method  & Venue & NM & BG & CL & Mean & Param. & MACs \\\midrule
GaitSet~\cite{chao2019gaitset} & AAAI19    &  95.0  &   87.2 &  70.4  &84.2      &  2.59         &  3.27          \\ 
GaitPart~\cite{Fan_2020_CVPR}&CVPR20        &   96.2   & 91.5   &  78.7  & 88.8   &  1.20    &   56.96                \\
GLN~\cite{hou2020gait}        &  ECCV20    & 96.9   &  94.0  &  77.5  &  89.5    &   14.70        &   22.14         \\
MT3D~\cite{lin2020gait}        &  MM20    &96.7    & 93.0   & 81.5   &  90.4    &   3.20        &    36.59        \\
CSTL~\cite{Huang_2021_ICCV}        &  ICCV21    &  97.8  & 93.6   &  84.2  &  91.9    &   9.09        &    6.43        \\
3DLocal~\cite{Huang_2021_ICCV2}        & ICCV21     &  97.5  &94.3    & 83.7   &  91.8    &   4.26        &   11.20         \\
GaitGL~\cite{Lin_2021_ICCV}        &  ICCV21    & 97.4   &  94.5  &  83.6  &  91.8    &    2.49       &  12.62          \\
GaitMPL~\cite{9769988}        &  TIP22    & 95.5   &  92.9  &  87.9  &  92.1    &    --       &  --          \\
Lagrange~\cite{Chai_2022_CVPR}        & CVPR22     &   96.9    &  93.5  &   86.5   &   92.3        &  --    &  --    \\ \midrule
 \rowcolor{lightgrey}\textbf{GaitGCI-T}      &  --    &  97.9  & 95.0   &  86.4  &  93.1    &    1.09       &  5.41          \\
 \rowcolor{lightgrey}\textbf{GaitGCI-M}       &   --   & 98.2   & 96.1   & 87.6   &  94.0    &     2.45      & 12.13           \\
 \rowcolor{lightgrey}\textbf{GaitGCI-L}        &   --   &  \textbf{98.4}  & \textbf{96.6} &  \textbf{88.5}  &   \textbf{94.5}   &  4.35         &   21.54        \\
\bottomrule
\end{tabular}
\label{tab:casia}
\end{table}

\begin{table}[htpb]
  \centering
  \small
  \renewcommand\arraystretch{0.8} 
  \setlength{\tabcolsep}{1.1mm}
  \caption{Rank-1 (\%), Rank-5 (\%), Rank-10 (\%), and Rank-20 (\%) performance comparison on GREW.}
\begin{tabular}{l|c|cccc}
\toprule
Method & Venue & Rank-1 & Rank-5 & Rank-10 & Rank-20 \\
\midrule
    PoseGait~\cite{liao2020model}& PR20& 0.2  & 1.1  & 2.2  & 4.8  \\ 
    GaitGraph~\cite{9506717}& ICIP21& 1.3  & 3.5  & 5.1  & 7.5  \\
    \midrule
    GEINet~\cite{7550060} &ICB16 & 6.8   & 13.4  & 17.0  & 21.0      \\
    TS-CNN~\cite{Wu2017}& TPAMI16& 13.6  & 24.6  & 30.2  & 37.0  \\ \midrule
    GaitSet~\cite{chao2019gaitset} & AAAI19& 46.3  & 63.6  & 70.3  & 76.8  \\
    GaitPart~\cite{Fan_2020_CVPR}& CVPR20& 44.0  & 60.7  & 67.3  & 73.5  \\
    GaitGL \cite{Lin_2021_ICCV} &ICCV21 & 47.3  & 63.6  & 69.3  &74.2  \\ \midrule
    \rowcolor{lightgrey}\textbf{GaitGCI}  & --& \textbf{68.5} & \textbf{80.8} & \textbf{84.9} & \textbf{87.7} \\
       \bottomrule
\end{tabular}
\label{tab:grew}
\end{table}
\noindent\textbf{GREW~\cite{zhu2021gait}.} GREW is one of the largest in-the-wild datasets, including 26345 subjects and 128671 sequences. It contains 4 modalities: silhouettes, optical flow, 2D/3D pose. GREW is divided into training set, validation set, and test set, containing 20000, 345, and 6000 subjects, respectively. During the evaluation, each subject contains 2 sequences as the probe and another 2 sequences as the gallery. 

\noindent\textbf{Gait3D~\cite{Zheng_2022_CVPR}.} Gait3D is the latest in-the-wild dataset containing 4000 subjects and 25309 sequences, which are collected in a large supermarket from 39 cameras. Following the protocol~\cite{Zheng_2022_CVPR}, 3000 subjects are selected as the training set, and the rest are the test set. For evaluation, one sequence of each subject is regarded as the query, and the other sequences become the gallery. Further, Gait3D provides 3D annotations to study model-based applications.

\subsection{Implementation Details} 

For common settings, the backbone is composed of 4 3D low-rank convolution layers. In the training stage, the frame number of each sequence is set to 30. The optimizer is Adam (lr=1$e$-4). The loss weight $\lambda$ is 0.1. The latent dimension $L$ and reduction ratio $r$ are set to 8 and 4, respectively. During the evaluation, all frames are fed into the framework. More details are in the supplementary material.

For CASIA-B, the channel $C$ of the backbone is set to (32, 64, 128, 128). We train the model for 80k iterations with batch size of (8,8). $M$ is set to 2. For other datasets, the network capacity should be increased~\cite{hou2020gait,Lin_2021_ICCV}. We add extra 2 layers with 128 channels. The batch size and $M$ are set to (32,8) and 8, respectively. The iterations are 200k, 200k, and 150k for OU-MVLP, GREW, and Gait3D, respectively.

\begin{table*}[t]
\small
\renewcommand{\arraystretch}{0.85}
\setlength{\tabcolsep}{3mm}
\centering
\caption{Rank-1 (\%), Rank-5 (\%), mAP (\%), and mINP (\%) comparison on Gait3D at the resolution of $128\times 88$ and $64\times 44$. As  skeleton-based methods are unrelated to the resolution, we only report one group of results. ``*'' denotes the method with extra 3D modality.}
\begin{tabular}{l|c|cccc|cccc}
\toprule
\multicolumn{2}{c|}{Input Size (H$\times$W)}  &\multicolumn{4}{c|}{128$\times$88}   & \multicolumn{4}{c}{64$\times$44} \\ \midrule
Methods                      			& Venue & Rank-1  & Rank-5  & mAP  & mINP & Rank-1 & Rank-5 & mAP  & mINP \\ \midrule
PoseGait~\cite{liao2020model}	        & PR20	& 0.2  & 1.1  & 0.5  & 0.3	& - & - & -  & - \\ 
GaitGraph~\cite{9506717}	& ICIP21& 6.3  & 16.2 & 5.2  & 2.4	& - & - & -  & - \\  \midrule
GaitSet~\cite{chao2019gaitset}		    & AAAI19	& 42.6 & 63.1 & 33.7 & 19.7	& 36.7 & 58.3 & 30.0 & 17.3	\\
GaitPart~\cite{Fan_2020_CVPR}		& CVPR20	& 29.9 & 50.6 & 23.3 & 13.2	& 28.2 & 47.6 & 21.6 & 12.4	\\ 
GLN~\cite{hou2020gait}				& ECCV20	& 42.2 & 64.5 & 33.1 & 19.6	& 31.4 & 52.9 & 24.7 & 13.6	\\
GaitGL~\cite{Lin_2021_ICCV}	            & ICCV21	& 23.5 & 38.5 & 16.4 & 9.2	& 29.7 & 48.5 & 22.3 & 13.3	\\ 
CSTL~\cite{Huang_2021_ICCV}	            & ICCV21	& 12.2 & 21.7 & 6.4  & 3.3	& 11.7 & 19.2 & 5.6  & 2.6	\\ \midrule

SMPLGait*~\cite{Zheng_2022_CVPR}& CVPR22  & 53.2	& 71.0 & 42.4 & 26.0 & 46.3& 64.5 & 37.2 & 22.2  \\  \midrule
\rowcolor{lightgrey}\textbf{GaitGCI}       & --  & \textbf{57.2}	& \textbf{74.5} & \textbf{45.0} & \textbf{27.6} & \textbf{50.3} & \textbf{68.5} & \textbf{39.5} & \textbf{24.3}  \\
\bottomrule
\end{tabular} 

\label{tab:gait3d}
\end{table*}

\subsection{Results under in-the-lab Scenario}

\noindent\textbf{OU-MVLP.} The comparison of~\cref{tab:ou} indicates that GaitGCI outperforms previous methods by a considerable margin, which reveals the effectiveness and generalizability of GaitGCI. In detail, GaitGCI achieves the best performance at almost all viewpoints. Specifically, performance at $0^\circ$/$180^\circ$ with less information is significantly improved, which may be attributed to reducing the impact of confounders so that the gait pattern is relatively salient.

\noindent\textbf{CASIA-B.} The comparison of~\cref{tab:casia} demonstrates that GaitGCI could efficiently outperform previous methods. Considering that GaitGCI is lightweight and increasing the number of channels could improve the network's capacity, we design three variants of GaitGCI, \ie., GaitGCI-T, GaitGCI-M, and GaitGCI-L with the channel $C$, $1.5C$, and $2C$, respectively. Specifically, GaitGCI-T could efficiently achieve 93.1\% rank-1 accuracy only with 1.09 M parameters and 5.41 G MACs. Further, GaitGCI-L could achieve 94.5\% rank-1 accuracy with acceptable costs. As a trade-off, GaitGCI-M could outperform GaitGL by 2.2\% with similar parameters and computation costs. Moreover, GaitGCI greatly improves the performance on BG/CL conditions, which suggests that confounders may hinder the development of existing methods on challenging conditions. The results of each view are in the supplementary material.

\subsection{Results under in-the-wild Scenario}

\noindent\textbf{GREW.} The performance comparison of skeleton-based, GEI-based, and silhouette-based methods on GREW is shown in~\cref{tab:grew}. Several conclusions could be drawn. First, the performance of the previous methods dramatically deteriorates when migrated to the in-the-wild scenario. Second, silhouette-based methods dominate the single-modality in-the-wild scenarios compared to skeleton/GEI-based methods. Third, GaitGCI significantly outperforms previous methods by over 20\%  and achieves $3^{rd}$ in the GREW competition~\cite{zhu2021gait} only using silhouette sequences. The results of GREW competition are in the supplementary material.
\begin{table}[t]
\centering
\renewcommand{\arraystretch}{0.85}
\caption{Ablation on counterfactual intervention learning (CIL) and diversity-constrained dynamic convolution (DCDC), which includes generative factual attention (GFA) and generative counterfactual attention (GCA).}
\begin{tabular}{ccc|ccc|c}
\toprule
CIL       & GFA & GCA & NM & BG & CL & Mean \\ \midrule
 &     &  & 96.5 & 92.9 & 80.9 & 90.1   \\
  \checkmark    &     &     & 97.1   &  93.8  & 84.2   &  91.7    \\
   \checkmark   & \checkmark    &     & 97.8   &  94.8   & 85.2   & 92.6     \\
   \checkmark  &     &\checkmark     & 97.7   & 94.5   &  85.3  &  92.5    \\
  \checkmark&   \checkmark  &   \checkmark  &  \textbf{97.9}  &  \textbf{95.0}  &  \textbf{86.4}  &  \textbf{93.1}   \\ \bottomrule
\end{tabular}
\label{tab:ab}
\end{table}
 \begin{figure}[t]
    \begin{center}
       \includegraphics[width=0.47\textwidth]{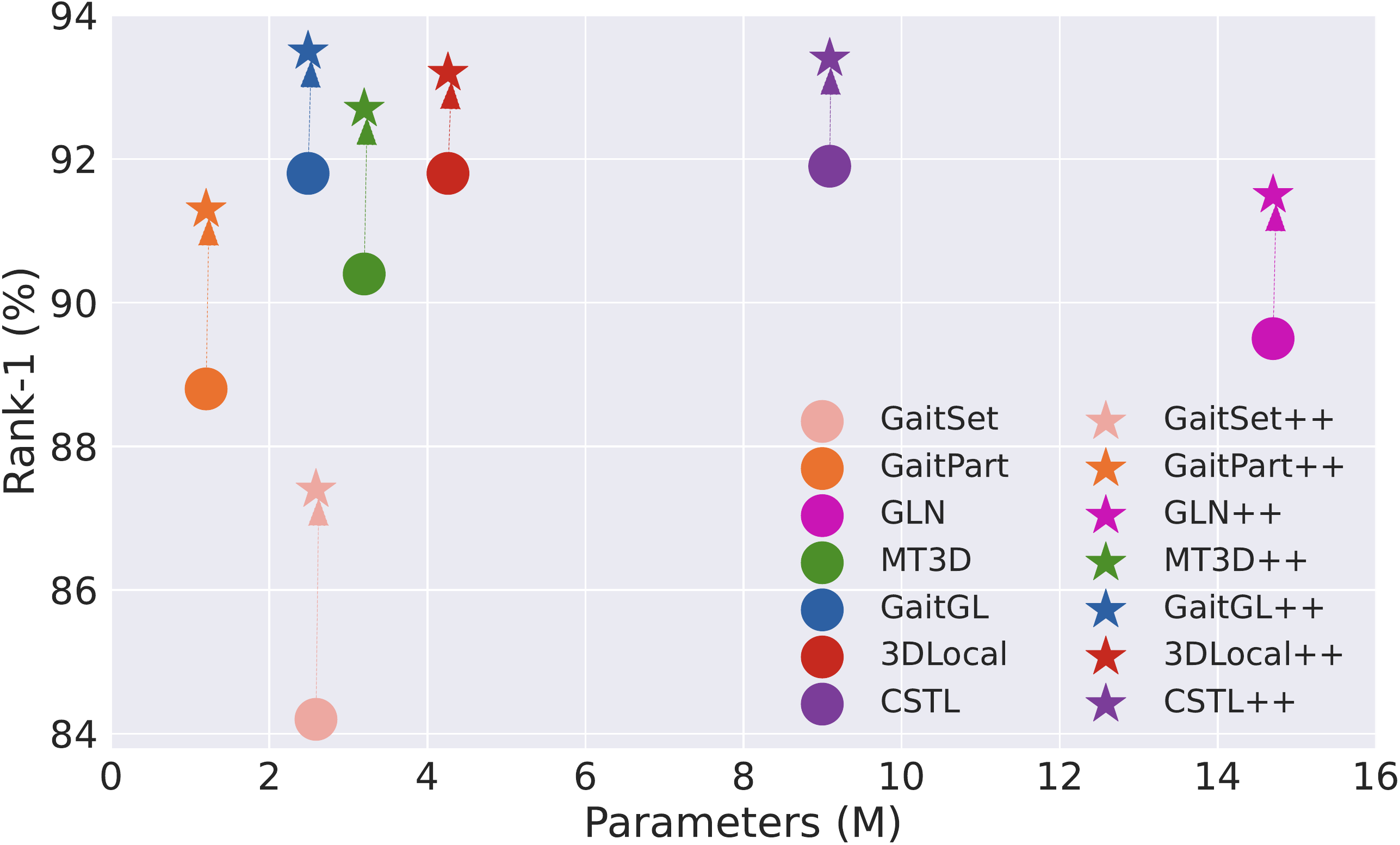}
    \end{center}
    \caption{Performance comparison of prevailing methods and those equipped with CIL and DCDC (denoted with suffix '++').}
    \label{fig:plus}
 \end{figure}

\noindent\textbf{Gait3D.} The comparison on the latest in-the-wild dataset Gait3D is conducted in~\cref{tab:gait3d}, including skeleton-based, silhouette-based, and multi-modal methods. GaitGCI outperforms prevailing silhouette-based methods by 14.6\% and 13.6\% in terms of rank-1 accuracy at the resolution of 128$\times$88 and 64$\times$44, respectively. Besides, the improvement of mAP and mINP fully illustrates the superior retrieval performance of GaitGCI. Further,  silhouette-based GaitGCI exceeds SMPLGait~\cite{Zheng_2022_CVPR}, which introduces extra 3D SMPL to perform multi-modal learning.

\noindent\textbf{Summary.} First, prevailing methods experience a dramatic performance decrease under in-the-wild scenarios, which indicates that the confounders under in-the-wild scenarios are more complex than those under in-the-lab scenarios. Second, the superior performance of GaitGCI under in-the-wild scenarios demonstrates the necessity for alleviating the impact of confounders. Third, although multi-modal methods dominate in-the-wild scenarios, silhouette-based methods have considerable performance improvement potential.

\begin{figure*}[ht]
\centering
	\subcaptionbox{Baseline on CASIA-B}{\includegraphics[width = 0.22\textwidth]{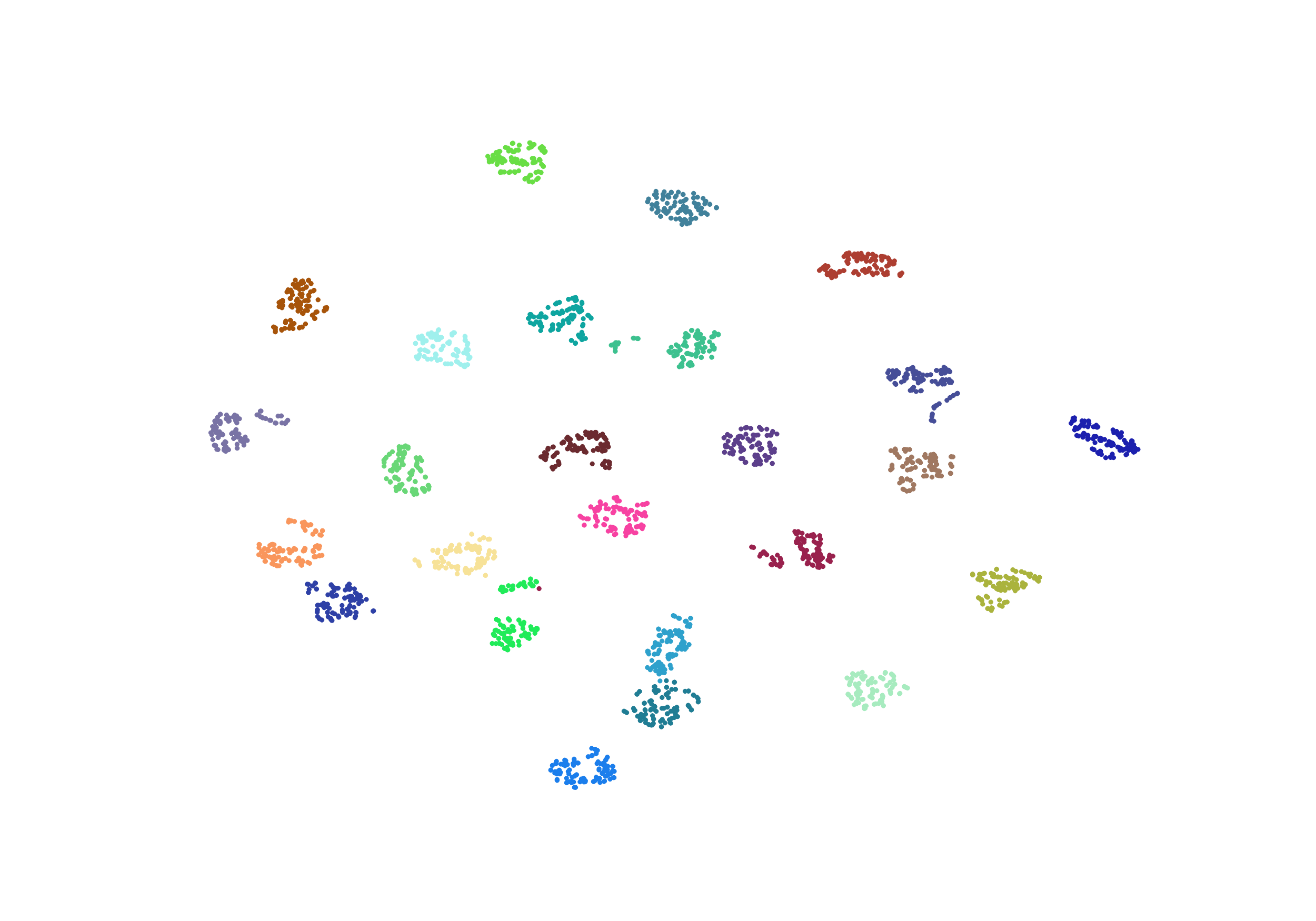}}
	\hspace{0.3cm}
	\subcaptionbox{GaitGCI on CASIA-B}{\includegraphics[width = 0.22\textwidth]{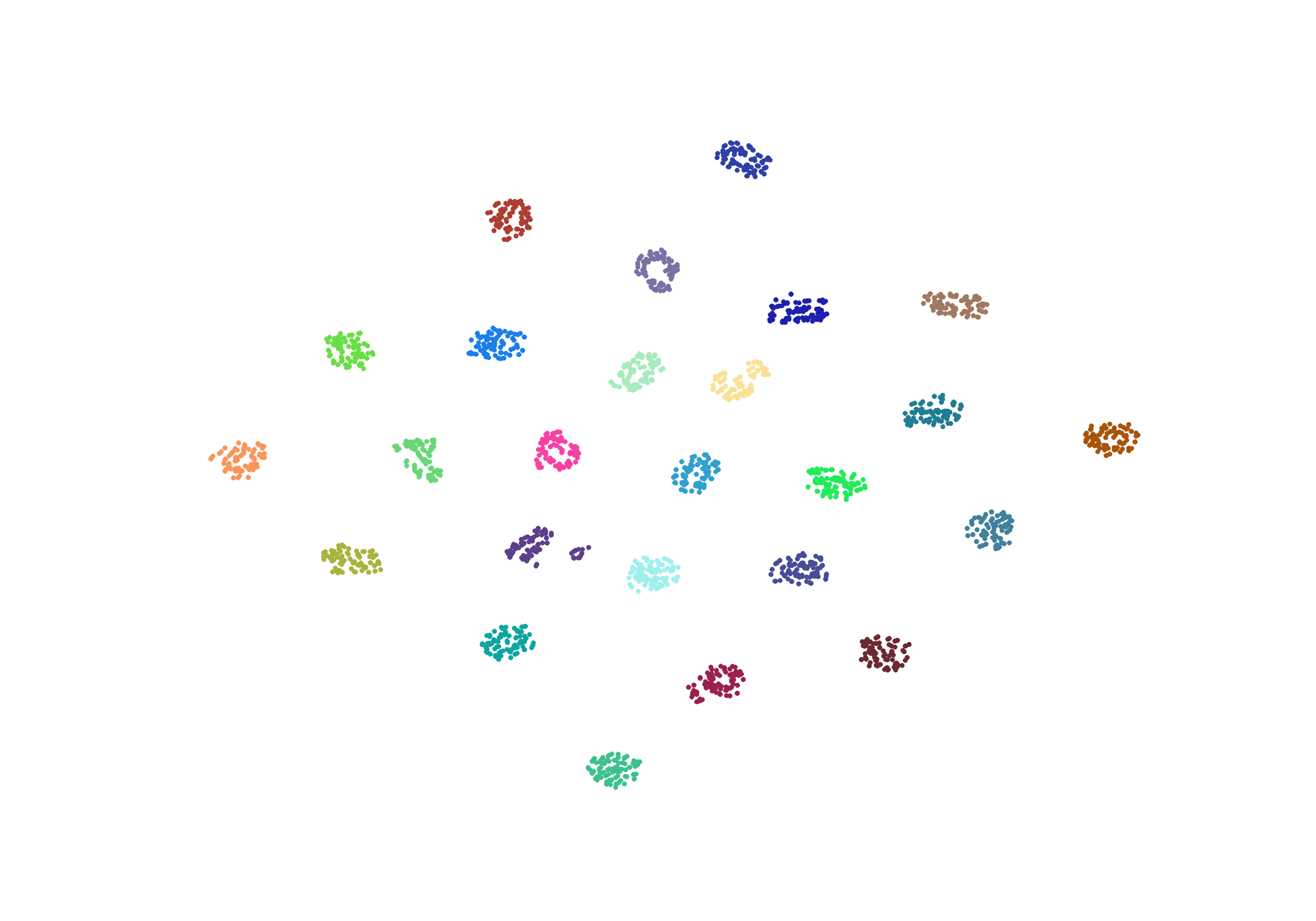}}
        \hspace{0.3cm}
        \subcaptionbox{Baseline on Gait3D}{\includegraphics[width = 0.22\textwidth]{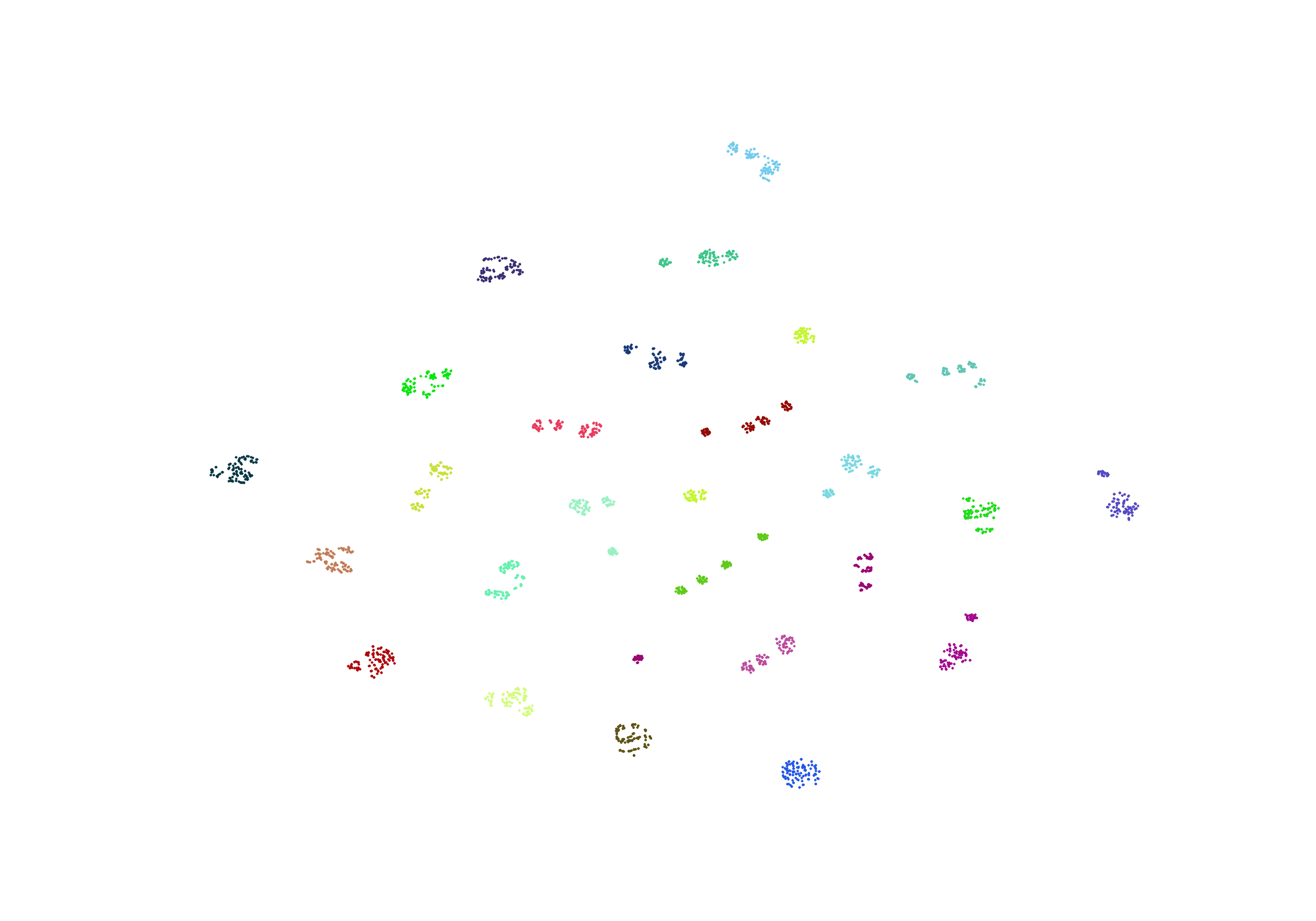}}
	\hspace{0.3cm}
	\subcaptionbox{GaitGCI on Gait3D}{\includegraphics[width = 0.22\textwidth]{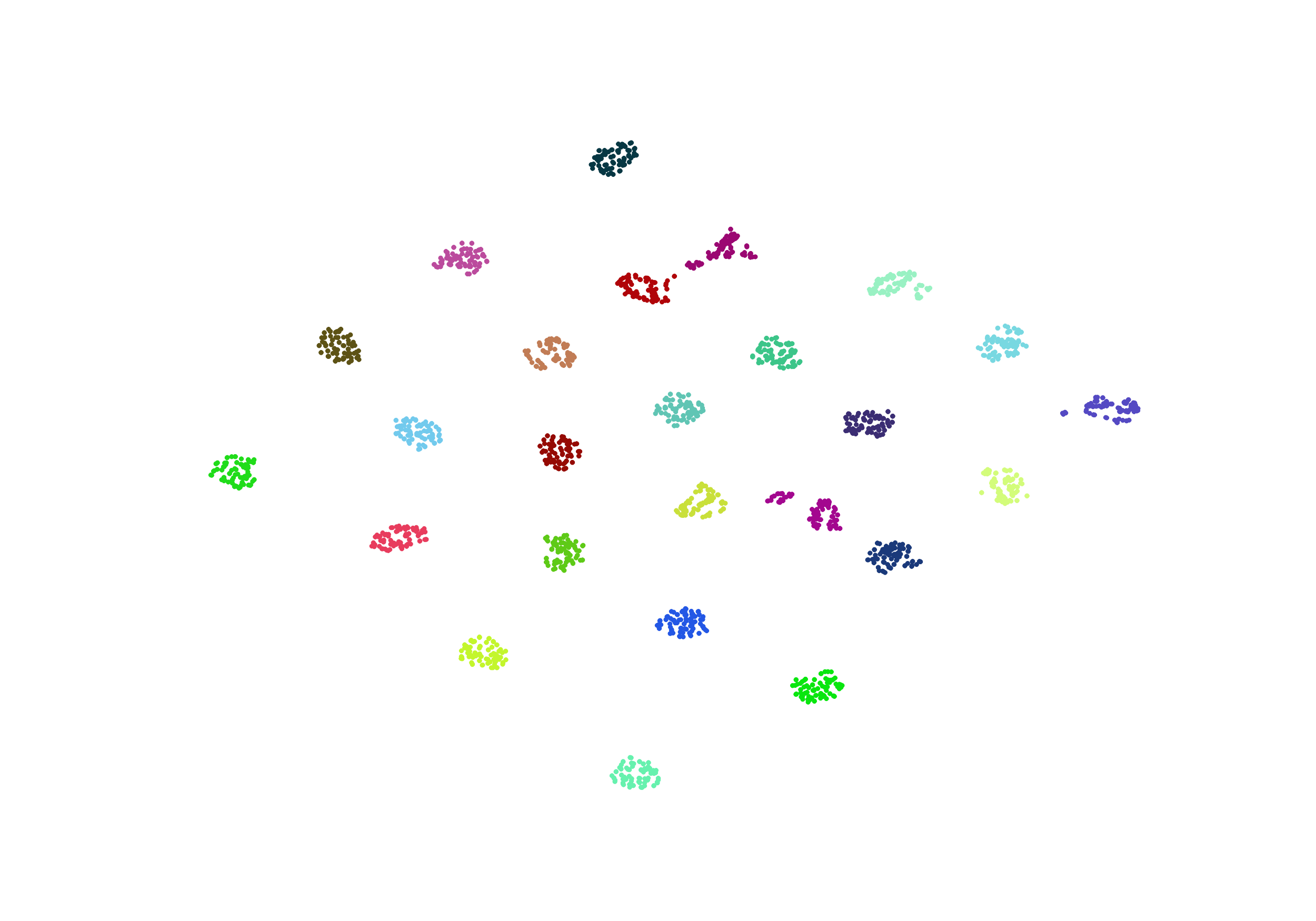}}
\caption{Comparison of the feature space under in-the-lab and in-the-wild scenarios using t-SNE~\cite{van2008visualizing}.}
\label{fig:sne}
\end{figure*}

\subsection{Ablation Study}
In this section, we conduct a series of quantitative and qualitative ablation studies to analyze the effectiveness of GaitGCI and its components. The baseline refers to the backbone with temporal pooling and separate FC~\cite{chao2019gaitset}.

\noindent\textbf{Individual Effectiveness of CIL and DCDC.}  The individual effects of CIL and DCDC are shown in~\cref{tab:ab}, where the factual/counterfactual attention of methods without GFA/GCA is set to static convolution and pre-defined normal distribution~\cite{9710619,chang2021towards}, respectively. CIL effectively improves 1.6\% rank-1 accuracy than baseline. Further, generative factual attention and generative counterfactual attention achieve 0.9\% and 0.8\% performance gain, respectively. And they deliver 1.4\% performance improvement in total, indicating the effectiveness and necessity of generative factual/counterfactual attention.

\noindent\textbf{Generalizability of GaitGCI.} As a model-agnostic module, CIL and DCDC could be plugged into prevailing methods. As shown in~\cref{fig:plus}, they could effectively boost the existing methods with nearly no extra costs, which indicates the generalizability and efficiency of CIL and DCDC. Further, this study demonstrates that the confounders may limit the performance of previous silhouette-based methods.

\begin{table}[t]
\centering
\caption{Analysis on DCDC. MD and DC denote matrix decomposition and diversity constraint, respectively.}
\renewcommand{\arraystretch}{0.8}
\begin{tabular}{l|cccc}
\toprule
Method     & NM & BG & CL & Mean \\ \midrule
Static Conv      &97.4&94.0&84.8&92.1\\ \midrule
DyConv    & 97.6    &94.4  &  85.8  &   92.6   \\
\hspace{1em}+MD   & 97.7   & 94.7   & 85.7   &   92.7   \\
\hspace{1em}+MD+DC & \textbf{97.9}   & \textbf{95.0}   &\textbf{86.4}    &  \textbf{93.1}  \\
\bottomrule
\end{tabular}
\label{tab:dc}
\end{table} 
\noindent\textbf{Analysis on DCDC.} To evaluate the effectiveness of diversity-constrained dynamic convolution on factual/counterfactual generation, the ablation is conducted in~\cref{tab:dc}. First, DyConv~\cite{chen2020dynamic} outperforms static convolution, which indicates the necessity of adaptiveness. Second, matrix decomposition could effectively reduce the computation and parameters while maintaining comparable performance, which demonstrates the redundancy of the high-dimensional computation of dynamic convolution. Third, the rank-based diversity constraint could efficiently improve the representation power.

\begin{figure}[t]
    \begin{center}
       \includegraphics[width=0.48\textwidth]{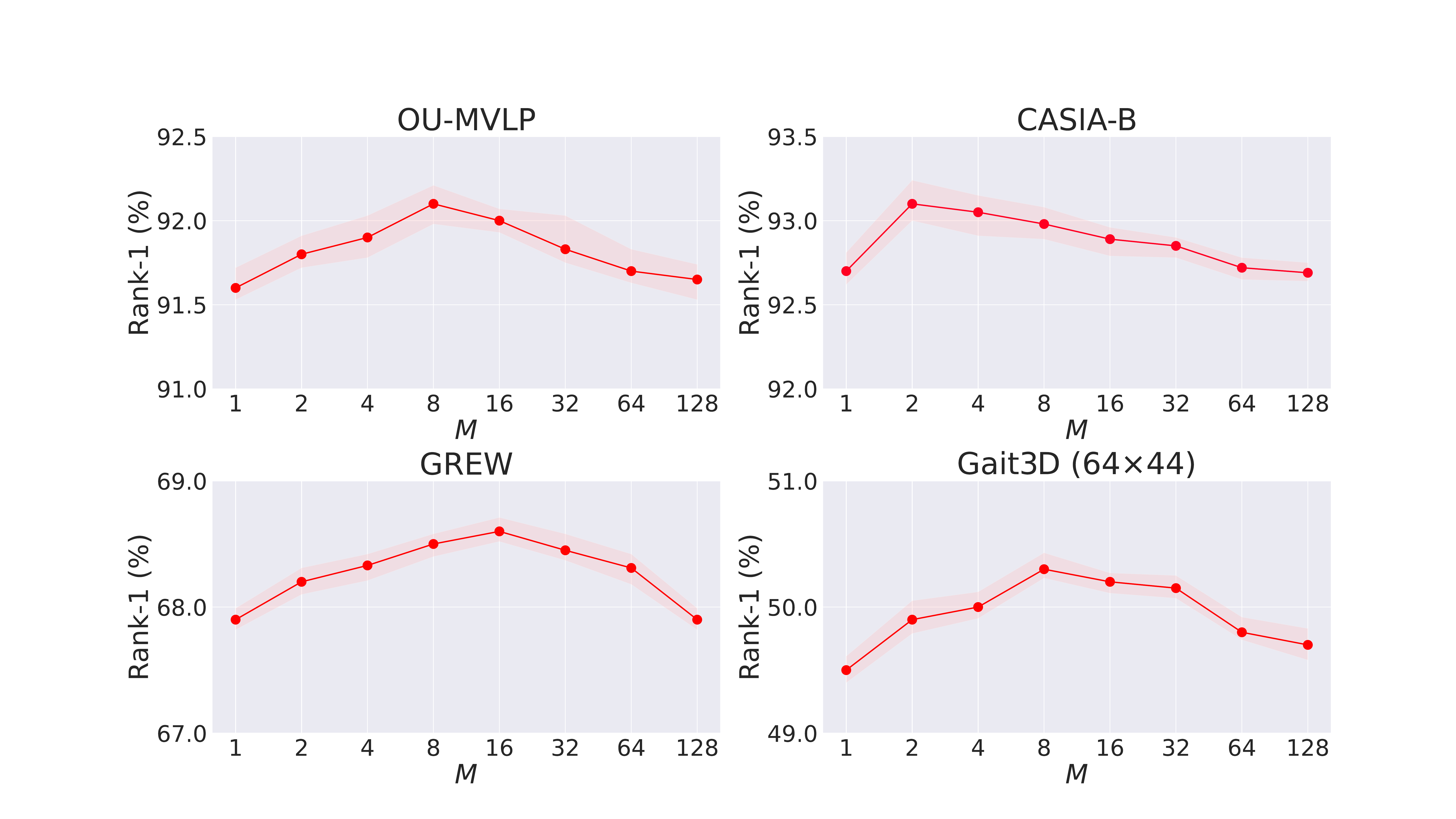}
    \end{center}
    \caption{Analysis of attention channel number $M$.}
    \label{fig:channel}
 \end{figure}

\noindent\textbf{Analysis on $M$.} The channel number $M$ controls the capacity to perceive sample-wise factual/counterfactual attention. From the results in~\cref{fig:channel}, we can conclude that: first, the in-the-wild dataset requires larger $M$, which may be due to the complexity of confounders and the dataset scale; second, the performance rises first and then falls with increasing $M$, which indicates that larger $M$ brings stronger capacity while superfluous $M$ may lead to overfitting.

\noindent\textbf{Visualization of Network Attention.} The visualization with Grad-CAM~\cite{selvaraju2017grad} is shown in~\cref{fig:gradcam}. Prevailing methods tend to collapse into confounders while neglecting most regions of the body boundary that could represent gait patterns. By alleviating the impact of confounders, GaitGCI could effectively focus on the discriminative and interpretable regions for gait pattern representation.

\noindent\textbf{Visualization of Feature Space.} To qualitatively evaluate the retrieval performance, we visualize the feature space by t-SNE~\cite{van2008visualizing} in~\cref{fig:sne}. First, GaitGCI could improve intra-class compactness and inter-class dispersibility under both scenarios. Second, the feature space of baseline under the in-the-lab scenario tends to have several sub-cluster in each cluster, and this phenomenon is more evident under the in-the-wild scenario, which may indicate the confounders of in-the-wild scenario are more complex. Meanwhile, it may also be why the previous model has acceptable performance under the in-the-lab scenario while the performance drops sharply under the in-the-wild scenario.

\section{Conclusion and Limitations}
This paper proposes a generative counterfactual intervention learning framework, which could force the network to focus on discriminative and interpretable regions. Counterfactual intervention learning leverages causal inference to analyze the direct causality link between factual attention and prediction. Further, diversity-constrained dynamic convolution, which could adaptively generate factual/counterfactual attention, utilizes matrix decomposition/diversity constraint to guarantee efficiency/representation power, respectively. Extensive experiments prove that GaitGCI could efficiently achieve state-of-the-art performance in arbitrary scenarios and could be used as a plug-and-play module.

For limitations, GaitGCI utilizes SVD, whose costs could only be ignored with low-dimensional feature representation. Besides, channel $M$ is a hyperparameter that depends on the dataset. In future work with high-dimensional representation and multi-dataset scenarios, we could alleviate these issues with numerical iteration methods~\cite{chen2019abd,bansal2018can} and attention-based channel selection, respectively.

\section*{Acknowledgements}
This work is supported in part by National Natural Science Foundation of China under Grant U20A20222, National Science Foundation for Distinguished Young Scholars under Grant 62225605, National Key Research and Development Program of China under Grant 2020AAA0107400, Zhejiang – Singapore Innovation and AI Joint Research Lab, Ant Group through CCF-Ant Research Fund, and sponsored by CCF-AFSG Research Fund, CAAI-HUAWEI MindSpore Open Fund as well as CCF-Zhipu AI Large Model Fund(CCF-Zhipu202302).

{\small
\bibliographystyle{ieee_fullname}
\bibliography{main}
}

\end{document}